\documentclass[11pt, a4paper, twocolumn]{article}

\usepackage{color}

\usepackage[utf8]{inputenc}
\usepackage[english]{babel}

\usepackage[top=2cm, bottom=2cm, left=1.9cm, right=1.9cm]{geometry}

\usepackage{titlesec}
\titlespacing{\section}{0pt}{\parskip}{\parskip}
\titlespacing{\subsection}{0pt}{\parskip}{\parskip}
\titlespacing{\subsubsection}{0pt}{\parskip}{\parskip}

\usepackage{indentfirst}
\usepackage[export]{adjustbox}
\usepackage{graphicx}
\usepackage{amssymb}

\usepackage[dvipsnames]{xcolor}
\graphicspath{{figures/}}

\usepackage[figurename=Fig.]{caption}
\usepackage{caption}
\usepackage[colorlinks, citecolor=green]{hyperref}

\usepackage{cite}

\usepackage{abstract}
\usepackage{multirow}
\usepackage{array}
\newcolumntype{*}{>{\global\let\currentrowstyle\relax}}
\newcolumntype{^}{>{\currentrowstyle}}
\newcommand{\rowstyle}[1]{\gdef\currentrowstyle{#1}%
	#1\ignorespaces
}

\title{\textbf{Learning Multi-scale Features for Foreground Segmentation}}
\date{\vspace*{2pt}}
\author{
	\normalsize \textbf{Long Ang Lim}\\
	\normalsize Ankara University\\
	\normalsize Department of Computer Engineering\\
	\normalsize lim.longang@gmail.com
	\and
	\normalsize \textbf{Hacer Yalim Keles}\\
	\normalsize Ankara University\\
	\normalsize Department of Computer Engineering\\
	\normalsize hkeles@ankara.edu.tr
}

\begin{document}
\maketitle

\begin{abstract}{
	\vspace*{-1.5em}
	\it Foreground segmentation algorithms aim segmenting moving objects from the background in a robust way under various challenging scenarios. Encoder-decoder type deep neural networks that are used in this domain recently perform impressive segmentation results. In this work, we propose a novel robust encoder-decoder structure neural network that can be trained end-to-end using only a few training examples. The proposed method extends the Feature Pooling Module (FPM) of FgSegNet by introducing features fusions inside this module, which is capable of extracting multi-scale features within images; resulting in a robust feature pooling against camera motion, which can alleviate the need of multi-scale inputs to the network. Our method outperforms all existing state-of-the-art methods in CDnet2014 dataset by an average overall F-Measure of 0.9847. We also evaluate the effectiveness of our method on SBI2015 and UCSD Background Subtraction datasets. The source code of the proposed method is made available at \href{https://github.com/lim-anggun/FgSegNet\_v2}{https://github.com/lim-anggun/FgSegNet\_v2}.
	
}\end{abstract}


\section{Introduction}
\label{sec:1}
	Extracting foreground objects from video sequences is one of the challenging and major tasks in computer vision domain. The resultant foreground objects, also known as moving objects, can be used in various computer vision tasks \cite{brutzer2011evaluation, porikli2003human, zhu2015human, poppe2010survey, cheung2004robust, basharat2008learning}. Extracting objects of interests from stationary camera-videos is challenging especially when the video sequences contain difficult scenarios such as sudden or gradual illumination changes, shadows, dynamic background motion, camera motion, camouflage or subtle regions. Several foreground segmentation approaches have been proposed to address these problems \cite{stauffer1999adaptive, zivkovic2004improved, barnich2011vibe, van2012background, kaewtrakulpong2002improved, bianco2017far, hofmann2012background} where most of the proposed methods rely on building the stationary background model; this approach is not very effective in adapting to the challenging scenarios.
	
	\begin{figure}[!t]
		\includegraphics[width=\columnwidth]{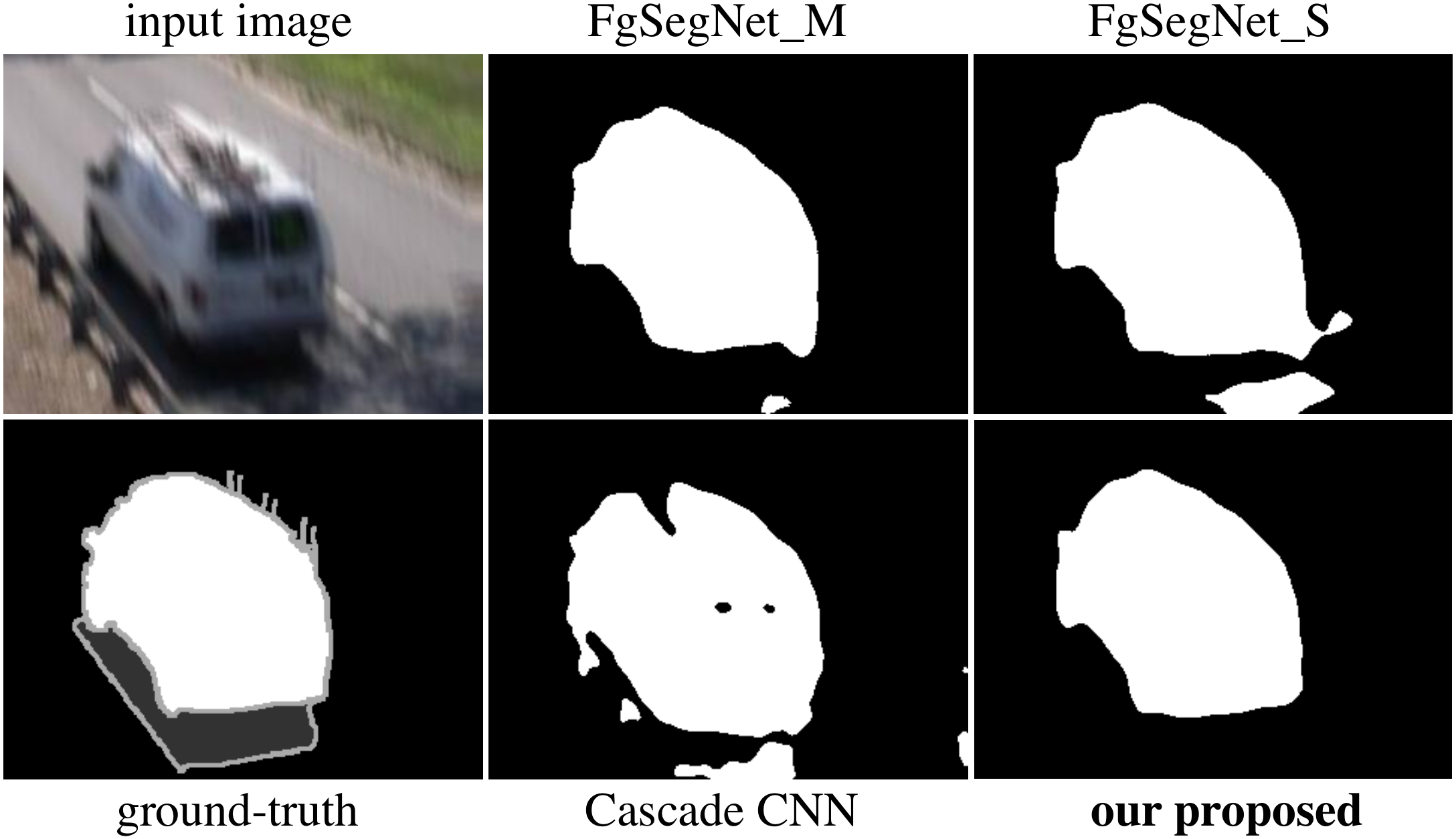}
		\caption{A comparison between our method and some current state-of-the-art methods on \textit{cameraJitter} category.}
		\centering
		\label{fig:fig1}
		\vspace*{-10pt}
	\end{figure}
	
	Convolutional Neural Networks (CNNs) \cite{lecun1998gradient} based on gradient learning are very powerful in extracting useful feature representations from data \cite{zeiler2014visualizing} and have been successfully used in many practical applications \cite{lim2018foreground, lecun1998gradient, krizhevsky2012imagenet, simonyan2014very, long2015fully, karpathy2015deep, badrinarayanan2015segnet, ronneberger2015u}. In particular, Fully Convolutional Networks (FCNs) that are based on transfer learning \cite{lim2018foreground, long2015fully} have shown significant improvement over conventional approaches by large margins. In this case, the knowledge that is gained from image classification problem is adapted to dense spatial class prediction domain where each pixel in an image is marked with a class label; such a prediction requires an understanding of both higher level and lower level contextual information in a scene. However, due to feature resolution reduction caused by consecutive pooling and strided convolution operations in the pre-trained models, this is usually difficult since the contextual details are lost. One may remove downsampling operations to keep high resolution feature maps, however it is computationally more expensive, and it is harder to expand the effective receptive fields. To take advantages of low level features in large resolution, \cite{lim2018foreground} recently removed the last block of VGG-16 \cite{simonyan2014very} and fine-tuned the last remaining block, and aggregated contextual information in multiple scales using a Feature Pooling Module (FPM).
	
	Motivated by the recent success of deep neural networks for foreground segmentation, we adapt the same encoder that is used in \cite{lim2018foreground}, which we found that it improves the performance more compared to other pre-trained networks, and we propose some modifications on the original FPM module to capture wide-range multi-scale information; resulting a more robust module against camera movements. In contrast to \cite{badrinarayanan2015segnet} that transfer max-pooling indices from the encoder to the decoder and \cite{ronneberger2015u} that copy feature maps directly from the encoder to the decoder to refine segmentation results, we use the global average pooling (GAP) from the encoder to guide the high level features in the decoder part.
	
	In summary, our key contributions are:
	\begin{itemize}
		\item We propose a robust foreground segmentation network that can be trained using only a few training examples without incorporating temporal data; yet, provided highly accurate segmentation results.
		\item We improve the FPM module by fusing multi-scale features inside that module; resulting in a robust feature pooling against camera motion, which can alleviate the need of multi-scale inputs to the network.
		\item We propose a novel decoder network, where high level features in the decoder part are guided by low level feature coefficients in the encoder part.
		\item Our method exceeds the state-of-the-art performances in Change Detection 2014 Challenge, SBI2015 and UCSD Background Subtraction datasets. 
		\item We provide ablation studies of design choices in this work and the source code is made publicly available to facilitate future researches.
	\end{itemize}
	
	
\section{Related Works}
\label{sec:2}
	Foreground segmentation, also known as background subtraction, is one of the major tasks in computer vision. Various methods have been proposed in this domain. Most conventional approach rely on building a background model for a specific video sequence. To model the background model, statistical (or parametric) methods using Gaussians are proposed \cite{stauffer1999adaptive, zivkovic2004improved, kaewtrakulpong2002improved} to model each pixel as a background or foreground pixel. However, parametric methods are computationally inefficient; to alleviate this problem, various non-parametric methods \cite{barnich2011vibe, van2012background, hofmann2012background} are proposed. Furthermore, \cite{bianco2017far} use Genetic programing to select different change detection methods and combined their results. 

	Recently, deep learning based methods \cite{lim2018foreground, WANG201766, sakkos2017end} show impressive results and outperform all classical approaches by large margins. There are different training strategies in this domain; for example, \cite{braham2016deep, babaee2017deep} use patch-wise training strategy where the background patches and the image patches are combined, and then fed to CNNs to predict foreground probabilities of the center pixels of the patches. However, this approach is computationally inefficient and may cause overfitting due to redundant pixels, loss of higher context information within patches, and requires large number of patches in training. \cite{lim2018foreground, WANG201766, sakkos2017end, lim2017background, cinelliforeground} approach the problem by using whole resolution images to the network to predict foreground masks. Some methods take advantages of temporal data \cite{sakkos2017end, lim2017background}, some methods train the networks by combining image frames with the generated background models \cite{braham2016deep, babaee2017deep, lim2017background, cinelliforeground}. The number of training data that is utilized to produce the model is also different in different approaches; \cite{braham2016deep}, \cite{babaee2017deep}, \cite{WANG201766}, \cite{lim2018foreground} and \cite{cinelliforeground} use 50\%, 5\%, 200 frames, 200 frames and 70\% from each video sequence, respectively; where \cite{sakkos2017end} split video sequences into chunks of 10 frames and use 70\% for training. In this research, we are generating a model for each scene and our purpose is to use only a few number of frames in training so that ground-truth requirement for different scenes will be very low. We believe as \cite{WANG201766, lim2018foreground} also do, this strategy is essential for a system that works in different domains in practice, since pixel-level ground-truth generation for large number of frames is a time-consuming and difficult process. Hence, we adapt the same training frames selection strategy as in \cite{WANG201766, lim2018foreground}, where only 200 frames or less are used for training. This strategy mitigates user interventions on labeling ground-truths considerably.
	\vspace{-16pt}
	\begin{figure*}[!th]
		\centering
		\includegraphics[width=0.95\textwidth]{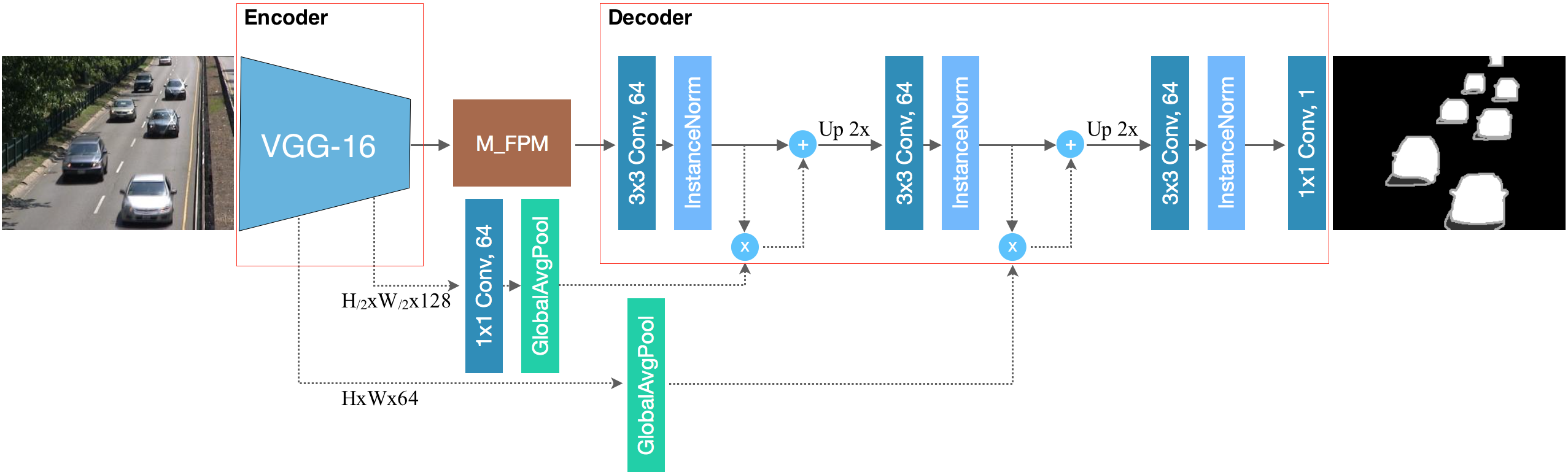}
		\caption{The flow of FgSegNet\_v2 architecture.}
		\label{fig:fig2}
	\end{figure*}
	
	\paragraph{Dilated Convolution:} Dilated convolution, also known as Atrous convolution, has been recently applied in semantic segmentation domain \cite{yu2015multi, chen2018deeplab, chen2017rethinking, chen2018encoder}; the idea is to enlarge the field-of-views in the network without increasing the number of learnt parameters. Motivated by the recent success of the previous works, we proposed in \cite{lim2018foreground} an FPM module with parallel dilated convolution layers that is plugged on top of a single-input encoder; it provides comparable foreground segmentation results compared to multi-inputs encoder.

\section{The Method}
\label{sec:3}
	In this section, we revisit our previous work, FgSegNet \cite{lim2018foreground}, in both encoder and FPM module. For more details, one may refer to the original paper.
	\subsection{The Encoder Network}
	\label{sec:3.1}
		Motivated by low level features of the pre-trained VGG-16 net \cite{simonyan2014very}, in FgSegNet, we utilize the first four blocks of VGG-16 net by removing the last, i.e. fifth, block and third max-pooling layer; we obtain higher resolution feature maps. Dropout \cite{srivastava2014dropout} layers are inserted after every convolutional layer of the modified net, and then this block is fine-tuned. In this work, we also use the same encoder architecture as in the FgSegNet implementation. We observe that this modified net improves the performance compared to other pre-trained nets.
		
	\subsection{The Modified FPM}
	\label{sec:3.2}
		\begin{figure}[!h]
			\centering
			\includegraphics[width=0.95\columnwidth, height=0.55\columnwidth]{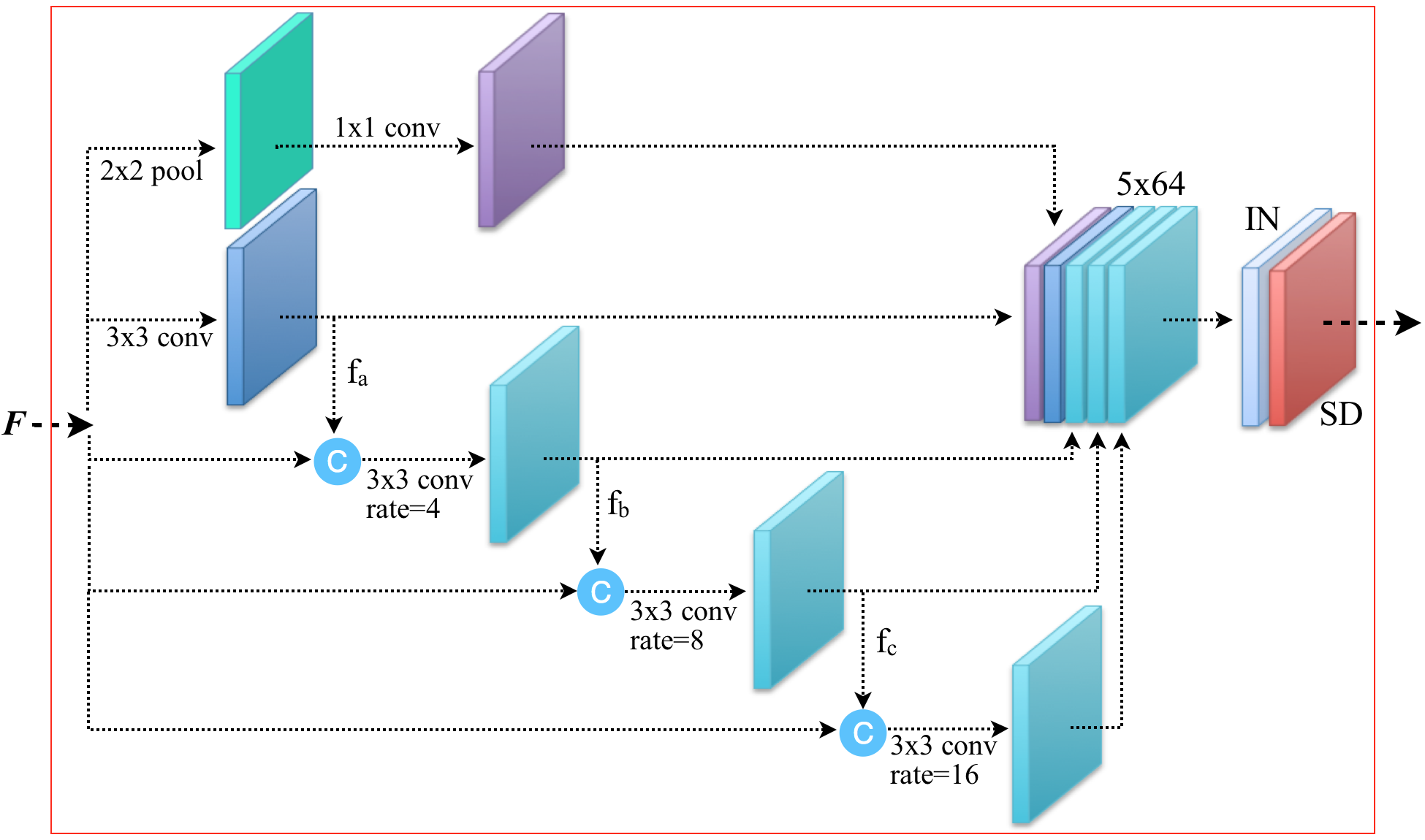}
			\caption{The modified FPM module, M-FPM. IN (InstanceNormalization), SD (SpatialDropout). All convolution layers have 64 features.}
			\label{fig:fig3}
		\end{figure}
		
		Given feature maps $F$, which are obtained from the output of our encoder, the original FPM module \cite{lim2018foreground} pools features at multiple scales by operating several convolutional layers with different dilation rates and a max-pooling layer followed by a 1x1 convolution on the same feature $F$, and then the pooled features are concatenated along the depth dimension. Finally, the concatenated feature is passed through BatchNormalization \cite{ioffe2015batch} and SpatialDropout \cite{tompson2015efficient} layers. In this work, we improve the original FPM module by proposing some modifications to it in two parts (Fig. \ref{fig:fig3}): (1) the resultant features $f_a$ from a normal 3x3-conv are concatenated with the feature $F$, and progressively pooled by a 3x3-conv with a dilation rate of 4, resulting in features $f_b$. Then, $F$ and $f_b$ are concatenated and fed to a 3x3-conv with dilation rate of 8, resulting in features $f_c$. Again, $F$ and $f_c$ are concatenated and fed to a 3x3-conv with dilation rate of 16. Finally, all features are concatenated to form 5x64 depth features, that we call it as $F’$; $F’$ contains multi-scale features with a wider receptive fields than the one in \cite{lim2018foreground}. (2) We replace BatchNormalization with InstanceNormalization \cite{Ulyanov2016InstanceNT} since we empirically observe that InstanceNormalization gives slightly better performance with a small batch size. Since multiple pooling layers are operated on the same features $F$, the concatenated features $F’$ are likely to be correlated. To promote independent feature maps, instead of the normal Dropout where individual neurons in a feature plane is dropped, SpatialDropout is used to drop the entire 2D feature maps by some rates, i.e. 0.25, if adjacent pixels within the feature maps are strongly correlated. We observe that SpatialDropout helps to improve the performance and prevents overfitting in our network. Note that we apply ReLU non-linearity once right after InstanceNormalization in the M-FPM module. We will refer to this modified FPM shortly as M-FPM from this on.
		
	\subsection{The Decoder with GAP Module}
	\label{sec:3.3}
		Our decoder network with two GAPs is illustrated in Fig. \ref{fig:fig2}. The decoder part contains the stack of three 3x3-conv layers and a 1x1-conv layer, where the 3x3-conv layers followed by InstanceNormalization and the 1x1-conv layer is the projection from feature space to image space. All 3x3-conv layers have 64 feature maps, except the 1x1-conv layer that contains 1 feature slice. Note that ReLU non-linearities are applied after InstanceNorm and sigmoid activation function is applied after 1x1-conv in the decoder part. 
		\vspace*{-20pt}

		\paragraph{Global Average Pooling (GAP):} There are two coefficients vectors that combine information from the low-level features of the encoder and high-level features of the decoder: the first coefficients vector is pooled from the second convolution layer right before max-pooling layer, the second coefficients vector is pooled from the fourth layer. Since 3x3-conv layers have 64 features in the decoder part, the fourth convolution layer of the encoder is first projected from 128 to 64 features. Both coefficient vectors (say $\alpha_i$) are multiplied with the output features of the first and second convolutional layers (say $f^i_j$) in the decoder part (see Fig. \ref{fig:fig2} for details). The scaled features are added with the original features to form features $f’^i_j$, where $f’^i_j:\alpha_i*f^i_j+f^i_j$, $i\in[0, 63]$ is the index of each feature depth and $j$ is the index of an element in each feature slice. Finally, the concatenated $f’^i_j$ are upscaled by 2x using bilinear interpolation and fed to the next layers. We observe that the network with GAP module add very slight computational cost but improves the performance in overall.
		
\section{Training Protocol}
\label{sec:4}
	We implement our models using Keras framework \cite{chollet2015keras} with Tensorflow backend with a single NVIDIA GTX 970 GPU. We follow the same training procedure as \cite{lim2018foreground}; hence, keeping the pre-trained coefficients of the original VGG-16 net, only the last modified block is fine-tuned. We train using \textit{RMSProp} optimizer (setting \textit{rho} to 0.9 and \textit{epsilon} to 1e-08) with a \textit{batch size} of 1. We use an initial learning rate of 1e-4 and reduce it by a factor of 10 when validation loss stops improving in 5 epochs. We set max 100 epochs and stop the training early when the validation loss stops improving in 10 epochs. The training frames (e.g. 200 frames) are hardly shuffled before \textit{training+validation} split and further split 80\% for training and 20\% for validation. The binary cross entropy loss is used to make an agreement between true labels and predicted labels. Due to highly imbalanced pixels between background/foreground pixels in scenes, we alleviate the imbalanced data classification problem by giving more weights to the rare class (foreground) but less weights to the major class (background) during training. Furthermore, since the output from the sigmoid activation are in range [0,1], we use them as the probability values; we apply thresholding to the activations to obtain discrete binary class labels as foreground and background.
	
	We mainly evaluate the model performance using F-Measure and Percentage of Wrong Classifications (PWC), where we want to maximize the F-Measure, while minimizing the PWC. Given true positive ($TP$), false positive ($FP$), false negative ($FN$), true negative ($TN$), F-Measure, is defined by:
	\begin{equation}
		F-Measure = \frac{2\times precision\times recall}{precision + recall}
	\end{equation}
	where $precision=\frac{TP}{TP + FP}$, $recall=\frac{TP}{TP + FN}$. And PWC is defined by:
	\begin{equation}
		PWC = \frac{100\times (FP + FN)}{TP + FP + TN + FN}
	\end{equation}
	\begin{table*}[!th]
		\caption{The results with GAP and no\_GAP}
		\label{table:1}
		\centering
		\scriptsize \begin{tabular}{*c^c^c^c^c|^c^c^c^c}
			\hline
			\multirow{3}{*}{Category} & \multicolumn{4}{c|}{no\_GAP} & \multicolumn{4}{c}{GAP} \\ \cline{2-9}
			&\multicolumn{2}{c}{F-Measure}&\multicolumn{2}{c|}{PWC}&\multicolumn{2}{c}{F-Measure}&\multicolumn{2}{c}{PWC} \\ \cline{2-9}
			&25f&200f&25f&200f&25f&200f&25f&200f \\
			\hline
			cameraJit&0.9506&0.9936&0.3026&0.0436&0.9740&0.9936&0.2271&0.0438 \\
			badWeather&0.9781&0.9858&0.0657&0.0271&0.9783&0.9848&0.0639&0.0295 \\
			dynamicBg&0.9636&0.9878&0.0311&0.0052&0.9665&0.9881&0.0325&0.0054 \\
			intermitt&0.9597&0.9929&0.2997&0.0794&0.9735&0.9935&0.3268&0.0707 \\
			shadow&0.9840&0.9960&0.1265&0.0279&0.9853&0.9959&0.1159&0.0290 \\
			turbulence&0.9600&0.9779&0.0439&0.0230&0.9587&0.9762&0.0438&0.0232 \\
			\hline
		\end{tabular}
	\end{table*}
\section{Results and Discussion}
\label{sec:5}
	In this section, we evaluate the effectiveness of our method using three different datasets, namely CDnet2014, SBI2015 and UCSD Background Subtraction. Each of these datasets contains challenging scenarios and used widely in fg/bg segmentation researches.
	
	\subsection{Experiments on CDnet Dataset}
	\label{sec:5.1}
		CDnet2014 dataset \cite{wang2014cdnet} contains 11 categories, where each category contains from 4 to 6 video sequences. Totally, there are 53 different video sequences in this dataset. The video sequence contains from 600 to 7999 frames with spatial resolutions varying from 320x240 to 720x576. Moreover, this dataset contains various challenging scenarios such as illumination change, hard shadow, highly dynamic background motion and camera motion etc.
		
		Unlike most previous works that use large numbers of training frames, to alleviate ground-truth labelling burden, \cite{lim2018foreground, WANG201766} use only a few frames, i.e. 50 and 200 frames, for \textit{training+validation}. Similar to these works, we use the same training set (i.e. 200 frames) provided by \cite{lim2018foreground} and we attempt to use a small number of training frames by randomly selecting 25 frames from the set of 200 frames to perform another experiment.
		\vspace{-10pt}
		
		\paragraph{The Global Average Pooling Experiments:} In order to evaluate the effectiveness of GAP layers in the proposed network, we performed two sets of experiments by selecting the most challenging 6 categories from the CDNet2014 dataset (\textit{cameraJitter, badWeather, dynamicBackground, intermittentObjectMotion, shadow, turbulence}); this subset of data totally contains 30 video sequences. The selected video sequences contain a number of frames that varies from 1150 to 7999. We use only 25 frames for \textit{training+validation} (as mentioned above) and the remaining frames for \textit{testing}. 
		
		In the first setting, we remove the GAP layers entirely from the network and will refer to this modified configuration as \textit{no\_GAP} below, and in the second setting we keep the GAP layers. As can be seen from Table \ref{table:1}, the network with GAP improves over \textit{no\_GAP} in most categories by some margins. Especially, GAP improves over \textit{no\_GAP} by 2.34\% points in \textit{cameraJitter} category.
		
		\begin{figure}[!h]
			\centering
			\includegraphics[width=\columnwidth]{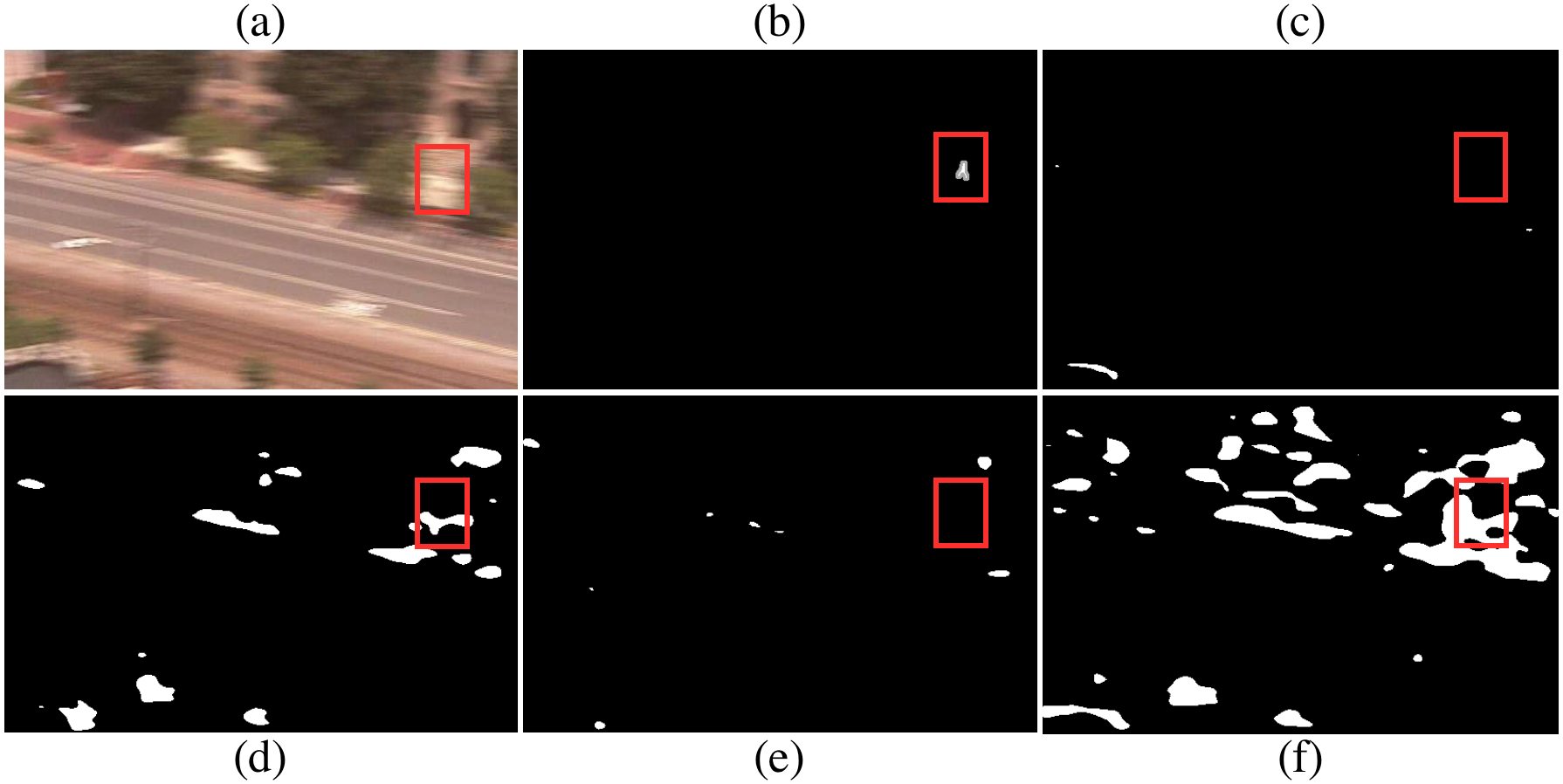}
			\caption{The improved M\_FPM module compared to the original FPM. (a) input image, (b) ground-truth, (c) M\_FPM+\textit{proposed decoder} result, (d) FPM+\textit{proposed decoder} result, (e) FgSegNet\_M result and (f) FgSegNet\_S result.}
			\label{fig:fig4}
		\end{figure}
	
		\begin{table*}[!th]
			\setlength{\tabcolsep}{4.5pt}
			\caption{The \textit{test results} are obtained by manually and randomly selecting, 25 and 200 frames from CDnet2014 dataset across 11 categories. Each row shows the average results of each category.}
			\label{table:2}
			\scriptsize \begin{tabular}{*c^c^c^c^c^c^c^c^c^c^c >{\bfseries}c>{\bfseries}c}
				\hline
				\multirow{2}{*}{Category} & \multicolumn{2}{c}{FPR} & \multicolumn{2}{c}{FNR} & \multicolumn{2}{c}{Recall} & \multicolumn{2}{c}{Precision} & 
				\multicolumn{2}{c}{PWC} & \multicolumn{2}{c}{\textbf{F-Measure}}\\\cline{2-13}
				&25f&200f&25f&200f&25f&200f&25f&200f&25f&200f&25f&200f \\
				\hline
				baseline&0.0002&0.00004&0.0100&0.0038&0.9900&0.9962&0.9942&0.9985&0.0480&0.0117&0.9921&0.9974 \\
				cameraJit&0.0004&0.00012&0.0419&0.0093&0.9581&0.9907&0.9907&0.9965&0.2271&0.0438&0.9740&0.9936 \\
				badWeather&0.0003&0.00009&0.0257&0.0215&0.9743&0.9785&0.9825&0.9911&0.0639&0.0295&0.9783&0.9848 \\
				dynamicBg&0.0001&0.00002&0.0315&0.0075&0.9685&0.9925&0.9655&0.9840&0.0325&0.0054&0.9665&0.9881 \\
				intermitt&0.0017&0.00015&0.0243&0.0104&0.9757&0.9896&0.9720&0.9976&0.3268&0.0707&0.9735&0.9935 \\
				lowFrameR.&0.0003&0.00008&0.2496&0.0956&0.7504&0.9044&0.7860&0.8782&0.1581&0.0299&0.7670&0.8897 \\
				nightVid.&0.0008&0.00022&0.1197&0.0363&0.8803&0.9637&0.9540&0.9861&0.3048&0.0802&0.9148&0.9747 \\
				PTZ&0.0002&0.00004&0.0870&0.0215&0.9130&0.9785&0.9776&0.9834&0.0892&0.0128&0.9423&0.9809 \\
				shadow&0.0003&0.0001&0.0203&0.0056&0.9797&0.9944&0.9911&0.9974&0.1159&0.0290&0.9853&0.9959 \\
				thermal&0.0009&0.00024&0.0456&0.0089&0.9544&0.9911&0.9815&0.9947&0.2471&0.0575&0.9677&0.9929 \\
				turbulence&0.0002&0.00011&0.0369&0.0221&0.9631&0.9779&0.9546&0.9747&0.0438&0.0232&0.9587&0.9762 \\
				\hline
				\rowstyle{\bfseries}Overall&0.0005&0.0001&0.0630&0.0220&0.9370&0.9780&0.9591&0.9802&0.1507&0.0358&0.9473&0.9789 \\
				\hline
			\end{tabular}
		\end{table*}
		\begin{table*}[!h]	
			\setlength{\tabcolsep}{2.5pt}
			\centering
			\caption{A comparison among 6 methods across 11 categories. Each row shows the results for each method. Each column shows the average results in each category. Note that we consider \textit{all the frames} in the ground-truths of CDnet2014 dataset.}
			\label{table:3}
			\scriptsize \begin{tabular}{*c^c^c^c^c^c^c^c^c^c^c^c^>{\bfseries}c}
				\hline
				\multirow{2}{*}{Methods}&\multicolumn{11}{c}{\textbf{F-Measure}}& \multirow{2}{*}{Overall} \\ \cline{2-12}
				&baseline&camJit&badWeat&dynaBg&intermit&lowFrame&nightVid&PTZ&shadow&thermal&turbul.&\\
				\hline
				\textbf{FgSegNet\_v2}&\textbf{0.9980}&\textbf{0.9961}&0.9900&0.9950&0.9939&\textbf{0.9579}&0.9816&\textbf{0.9936}&0.9966&0.9942&\textbf{0.9815}&0.9890 \\
				FgSegNet\_S \cite{lim2018foreground}&\textbf{0.9980}&0.9951&\textbf{0.9902}&\textbf{0.9952}&\textbf{0.9942}&0.9511&\textbf{0.9837}&0.9880&\textbf{0.9967}&\textbf{0.9945}&0.9796&0.9878 \\
				FgSegNet\_M \cite{lim2018foreground}&0.9975&0.9945&0.9838&0.9939&0.9933&0.9558&0.9779&0.9893&0.9954&0.9923&0.9776&0.9865 \\
				Cascade CNN \cite{WANG201766}&0.9786&0.9758&0.9451&0.9658&0.8505&0.8804&0.8926&0.9344&0.9593&0.8958&0.9215&0.9272 \\
				DeepBS \cite{babaee2017deep}&0.9580&0.8990&0.8647&0.8761&0.6097&0.5900&0.6359&0.3306&0.9304&0.7583&0.8993&0.7593 \\
				IUTIS-5 \cite{bianco2017far}&0.9567&0.8332&0.8289&0.8902&0.7296&0.7911&0.5132&0.4703&0.9084&0.8303&0.8507&0.7820 \\
				\hline
			\end{tabular}
		\end{table*}
	
		\paragraph{The Modified FPM (M-FPM) Experiments:} In this study, we demonstrate the effectiveness of the M-FPM module compared to the original FPM proposed by \cite{lim2018foreground}. We again perform two set of experiments; in the first setting, we utilize the proposed decoder with the M-FPM module, while in the second, we use the proposed decoder with original FPM module. The experimental results are illustrated using a challenging scene,in Fig. \ref{fig:fig4}, where previous networks produce many false positives. As can be seen, (1) the proposed M-FPM module (Fig. \ref{fig:fig4}, (c)) produces less false positive compared to the original FPM module (Fig. \ref{fig:fig4}, (d)), (2) the proposed decoder (Fig. \ref{fig:fig4}, (d)) is effective compared to the FgSegNet family \cite{lim2018foreground} decoder (Fig. \ref{fig:fig4}, (f)), (3) since \textit{FgSegNet\_M} \cite{lim2018foreground} fuses and jointly learns multi-inputs network features, it is robust to camera movement (Fig. \ref{fig:fig4}, (e)). As a comparison, we empirically observe that the M-FPM module can mitigate the need of multi-inputs network, which is computationally more expensive, by introducing the multi-scale feature fusion later instead; resulting in a robust feature pooling module. As can be seen from Fig. \ref{fig:fig4}, the proposed method (Fig. \ref{fig:fig4}, (c)) produces very less false positives compared to FgSegNet family (Fig. \ref{fig:fig4}, (e) and (f)) and improves over \textit{FgSegNet\_M}, \textit{FgSegNet\_S} and \textit{Cascade CNN} \cite{WANG201766} by 0.43\%, 0.56\% and 5.92\% points, respectively, in \textit{PTZ} category (see Table \ref{table:3}). We also show the effectiveness of proposed M-FPM in Fig. \ref{fig:fig1}.
	
		After evaluating the effectiveness of the extensions that we propose in this work with the challenging subset of CDnet2014 dataset, we perform further experiments by using the proposed architecture configuration (Fig. \ref{fig:fig2}). As mentioned above, labeling the dense ground-truths requires more human-efforts and to reduce labelling burden, \cite{lim2018foreground, WANG201766} use only a few training examples. We adapt the same idea for 200-frames experiment; however, in this work, we further reduce the training examples by 8x to 25 frames. Specifically, we take two sets of experiments by using 25 frames and 200 frames and illustrate the \textit{test results} in Table \ref{table:2}. As can be seen, for 25-frames experiments, we obtain an overall F-Measure of 0.9473 across 11 categories. By increasing the number of frames to 200, F-Measure increases by 3.16\% compared to the 25-frame experiment results. Similarly, the PWC decreases by 0.115\% when we increase the number of frames from 25 to 200. Note that the training frames are not included in these evaluations, only the \textit{test frames} are utilized. We use threshold of 0.7 for 25-frames experiments and threshold of 0.9 for 200-frames experiment since the network provides best performances with these settings.
		
		\begin{table}[!h]			
			\setlength{\tabcolsep}{3.5pt}
			\centering
			\caption{A comparison with the state-of-the-art methods. These average results are obtained from Change Detection 2014 Challenge.}
			\label{table:4}
			\scriptsize\begin{tabular}{*c^c^c^c^c}
				\hline
				\multirow{2}{*}{Methods}&\multicolumn{4}{c}{\textbf{Overall}} \\ \cline{2-5}
				&Precision&Recall&PWC&F-Measure \\
				\hline
				\rowstyle{\bfseries}FgSegNet\_v2&0.9823&0.9891&0.0402&0.9847 \\
				FgSegNet\_S \cite{lim2018foreground}&0.9751&0.9896&0.0461&0.9804 \\
				FgSegNet\_M \cite{lim2018foreground}&0.9758&0.9836&0.0559&0.9770 \\
				Cascade CNN \cite{WANG201766}&0.8997&0.9506&0.4052&0.9209 \\
				DeepBS \cite{babaee2017deep}&0.8332&0.7545&1.9920&0.7458 \\
				\hline
				IUTIS-5 \cite{bianco2017far}&0.8087&0.7849&1.1986&0.7717 \\
				\hline
			\end{tabular}
			\vspace*{-7pt}
		\end{table}
	
		We further compare the results between the proposed method and state-of-the-art methods in Table \ref{table:3}. Note that to make a comparison in terms of the number of frames (follow \textit{changedetection.net}), we tested our model using all the provided ground-truths in CDnet2014 dataset. As can be seen, our method (\textit{FgSegNet\_v2}) improves over current state-of-the-art methods by some margins. In particular, it significantly improves on camera motion categories (as discussed above); i.e. \textit{PTZ} and \textit{cameraJitter} category, where camera is shaking, panning, tilting or zooming around the scenes. This eliminates the tradeoff of using multi-inputs features fusion in \textit{FgSegNet\_M}.
	
		\begin{figure*}[h]
			\centering
			\includegraphics[width=\textwidth]{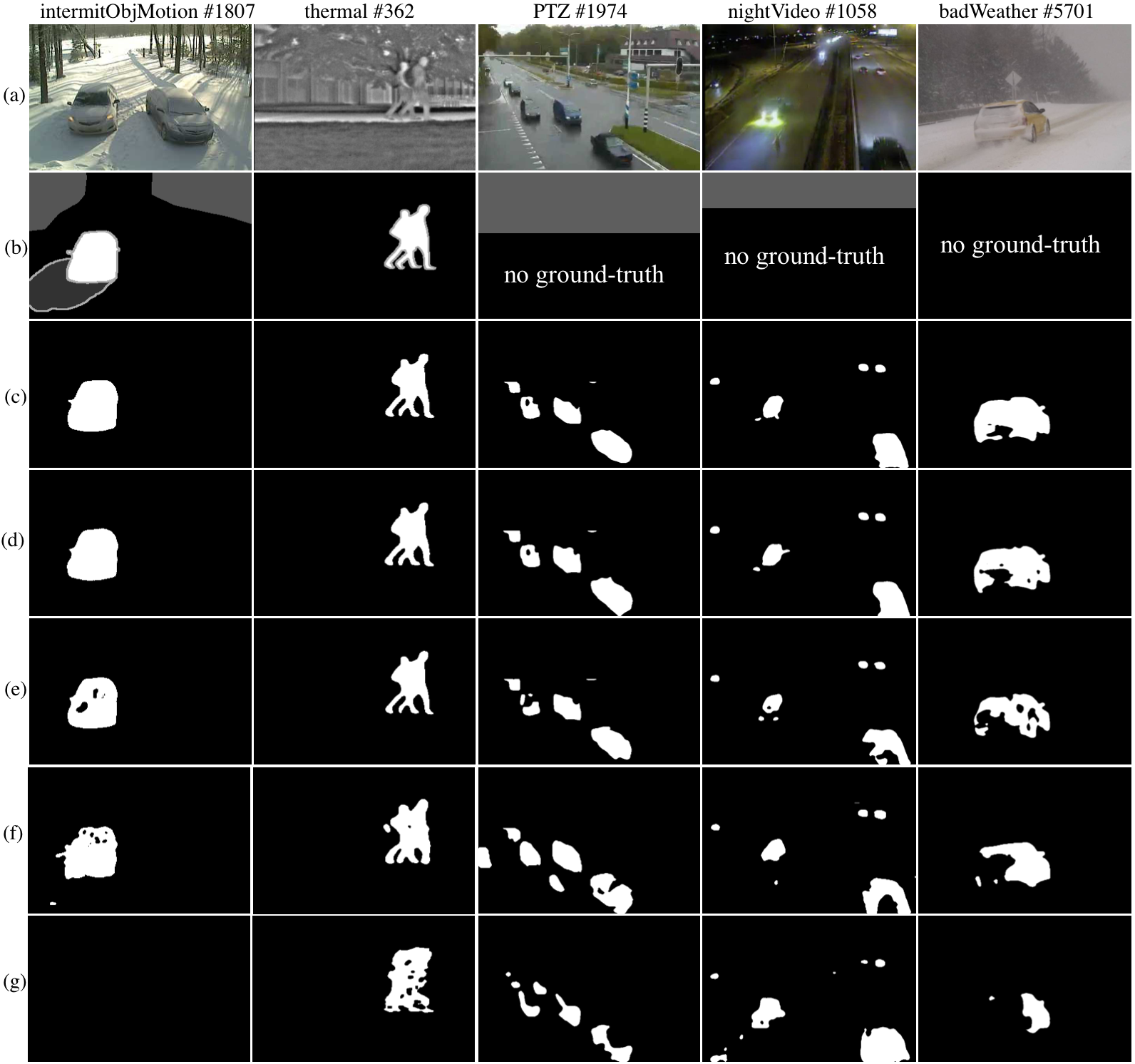}
			\caption{Some comparisons among 5 methods. (a) input images, (b) ground-truths, (c)  our segmentation results, (d) FgSegNet\_S results, (e) FgSegNet\_M results, (f) Cascade CNN results and (g) DeepBS results. \# (frame number)}
			\label{fig:fig5}
		\end{figure*}
	
		We further perform another experiment by using the training frames provided by \cite{WANG201766} and evaluate our model on Change Detection 2014 Challenge (changedetection.net). The comparison is provided in Table \ref{table:4}. As can be seen, our method outperformed existing state-of-the-art methods by some margins; specifically, for the deep learning based methods, it improves over \textit{FgSegNet\_S}, \textit{FgSegNet\_M}, \textit{Cascade CNN} and \textit{DeepBS} by 0.43\%, 0.77\%, 6.38\% and 23.89\% points, respectively. It also improves over all traditional methods over 21.3\% points. Our method is ranked as number 1 at the time of submission.
	
		In order to display the generalization capability of all methods with some exemplary frames from the dataset, we provide segmentation results in two ways; first, we choose some segmentation results randomly in the range of the provided ground-truths (i.e. \textit{intermitObjMotion} and \textit{thermal} categories), second, we randomly chose some example frames from the \textit{test set} where ground-truths are not publicly shared (i.e. \textit{PTZ, nightVideo} and \textit{badWeather} categories). As can be seen from Fig. \ref{fig:fig5}, our method gives good segmentation results; especially, in \textit{intermitObjMotion} category where a car stopped in the scene for a long time and starts moving immediately. This is the case where most methods fail dramatically in this scenario (e.g. method in (g)). As for \textit{test results} where ground-truths are not available in Fig. \ref{fig:fig5} (\textit{PTZ, nightVideo, badWeather} categories), our method generalizes well to completely unseen data where challenging content of the scenes change over time. Our method produces good segmentation masks compared to other methods.
	
\begin{figure}[!h]
			\centering
			\includegraphics[width=\columnwidth]{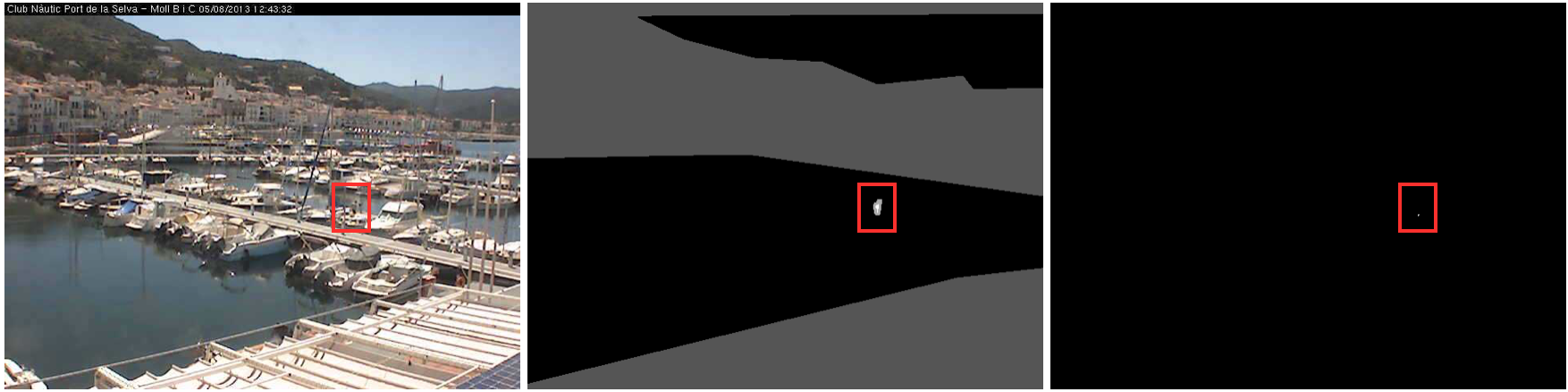}
			\caption{The video sequence in \textit{lowFrameRate} category that our method performs poorly. First, second and third column show input image, ground-truth and our segmentation result, respectively.}
			\label{fig:fig6}
		\end{figure}

		Our method performs better in almost all categories, except \textit{LowFrameRate} category where it performs poorly (F-Measure=0.9579) compared to other categories (see Table \ref{table:3}). This low performance is primarily due to a challenging video sequence (in \textit{lowFrameRate} category), where there are extremely small foreground objects in dynamic scenes with gradual illumination changes (Fig. \ref{fig:fig6}). In this case, the network may pay more attention to the major class (bg) but less attention to the rare class (fg); resulting in misclassifying very small foreground objects. However, the proposed method still improves over the best method by some margins in this category. Furthermore, our method fails to detect blended objects into the scene (see Fig. \ref{fig:fig4}); in this case, an object is blended to the background completely and it is even hard for human to distinguish between foreground and background.
		
	\subsection{Experiments on SBI2015 Dataset}
	\label{sec:5.2}
		We conduct additional experiments on Scene Background Initialization 2015 (SBI2015) dataset \cite{maddalena2015towards}, which contains 14 video sequences with ground-truth labels provided by \cite{WANG201766}. We follow the same training protocol as in \cite{lim2018foreground, WANG201766}, where 20\% of the frames are used for \textit{training+validation} and the remaining 80\% split for testing; if we denote the number of training examples used in training as $t_n$, in these experiments $t_n \in [2, 148]$. 
		
\begin{figure*}[!h]
			\centering
			\includegraphics[width=0.98\textwidth]{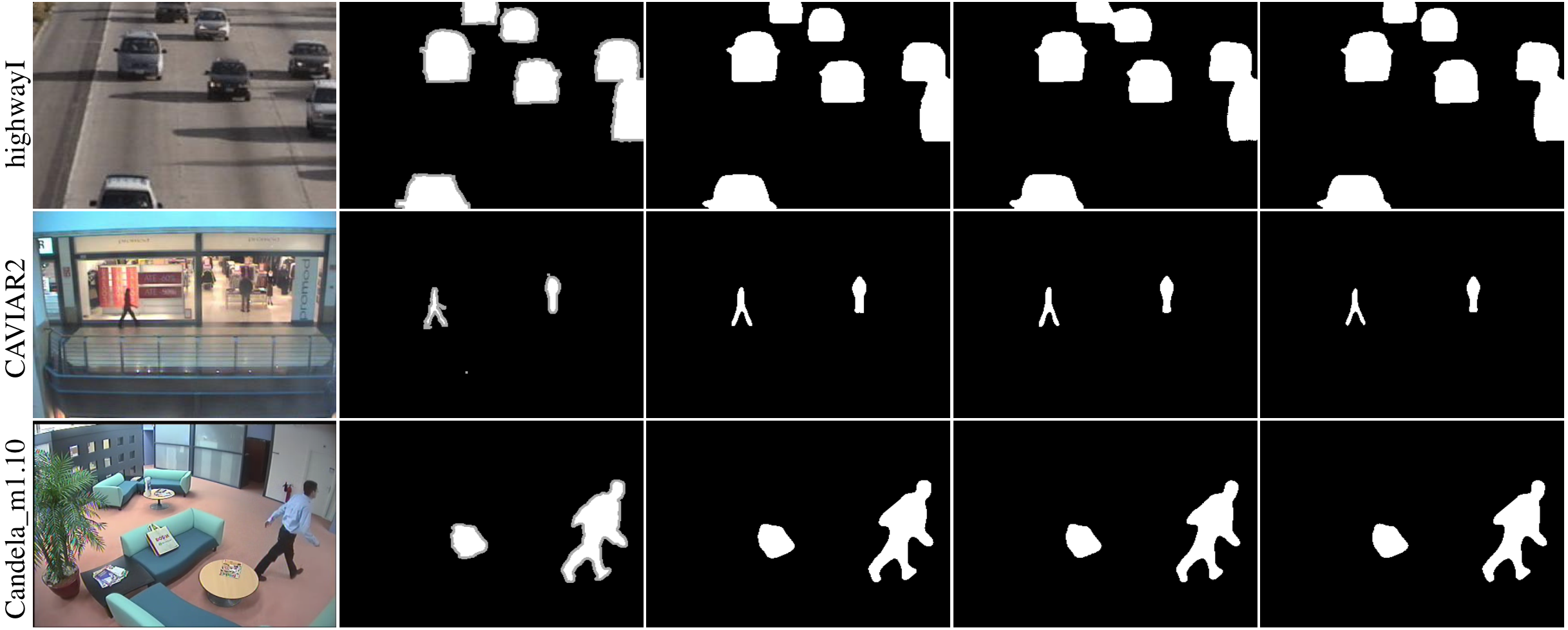}
			\caption{A comparison on SBI2015 dataset. Each column shows raw images, the ground-truths, our segmentation results, FgSegNet\_S and FgSegNet\_M results, respectively.}
			\label{fig:fig7}
		\end{figure*}

		The \textit{test results} are illustrated in Table \ref{table:5}. As can be seen, our method outperformed previous state-of-the-art methods by some margins. To be concrete, it improves over FgSegNet family by 0.22\% and 0.59\% points, while improves over \textit{Cascade CNN} \cite{WANG201766} by 9.21\% points. Similarly, the PWC of our method is significantly less compared to other methods. We obtain the lowest performance in \textit{Toscana} sequence with an F-Measure of 0.9291; this is primarily due to using very low number of training frames; only 2 frames, for \textit{training+validation}. We depict some exemplary results in Fig. \ref{fig:fig7}. As can be seen, our method produces good segmentation masks; especially, in \textit{highwayI} video sequence where hard shadows are eliminated completely.
		\begin{table}[!h]			
			\setlength{\tabcolsep}{1.5pt}
			\renewcommand{\arraystretch}{1.2}
			\centering
			\caption{The \textit{test results} on SBI2015 dataset with threshold of 0.3 and comparisons with the state-of-the-art methods.}
			\label{table:5}
			\scriptsize 
			\begin{tabular}{*c^c^c^c^c}
				\hline
				Video Seq.&FPR&FNR&F-Measure&PWC \\
				\hline
				Board&0.0009&0.0019&0.9979&0.1213 \\
				Candela\_m1.10&0.0003&0.0037&0.9950&0.0399 \\
				CAVIAR1&0.0001&0.0007&0.9988&0.0086 \\
				CAVIAR2&0.0001&0.0092&0.9834&0.0131 \\
				CaVignal&0.0027&0.0076&0.9859&0.3310 \\
				Foliage&0.0771&0.0207&0.9732&3.7675 \\
				HallAndMonitor&0.0002&0.0051&0.9926&0.0357 \\
				HighwayI&0.0008&0.0068&0.9924&0.1358 \\
				HighwayII&0.0001&0.0051&0.9952&0.0289 \\
				HumanBody2&0.0009&0.0079&0.9920&0.1636 \\
				IBMtest2&0.0007&0.0220&0.9817&0.1680 \\
				PeopleAndFoliage&0.0066&0.0102&0.9919&0.8468 \\
				Snellen&0.0211&0.0147&0.9644&1.7573 \\
				Toscana&0.0046&0.1155&0.9291&2.5901 \\
				\hline
				\rowstyle{\bfseries}FgSegNet\_v2&0.0083&0.0165&0.9853&0.7148 \\
				\hline
				\rowstyle{\bfseries}FgSegNet\_S \cite{lim2018foreground}&0.0090&0.0146&0.9831&0.8524 \\
				\rowstyle{\bfseries}FgSegNet\_M \cite{lim2018foreground}&0.0059&0.0310&0.9794&0.9431 \\
				\rowstyle{\bfseries}Cascade CNN \cite{WANG201766}&--&--&0.8932&5.5800 \\
				\hline
			\end{tabular}
		\end{table}
	
	\subsection{Experiments on UCSD Dataset}
	\label{sec:5.3}
		\begin{table*}[!th]			
			\centering
			\caption{The \textit{test results} on UCSD dataset with threshold of 0.6 and some comparisons with state-of-the-art methods.}
			\label{table:6}
			\scriptsize 
			\begin{tabular}{*c^c^c^c^c^c^c^c^c }
				\hline
				\multirow{2}{*}{Video Seq.}&\multicolumn{4}{c}{\textbf{20\% split}}&\multicolumn{4}{c}{\textbf{50\% split}} \\ \cline{2-9}
				&FPR & FNR & F-Measure & PWC &FPR & FNR & F-Measure & PWC\\
				\hline
				birds&0.0025&0.1423&0.8649&0.5205&0.0021&0.1162&0.8884&0.4315\\
				boats&0.0009&0.0729&0.9213&0.1678&0.0008&0.0382&0.9437&0.1213\\
				bottle&0.0014&0.0406&0.9550&0.2462&0.0009&0.0476&0.9605&0.2134\\
				chopper&0.0023&0.0760&0.9140&0.3991&0.0017&0.0810&0.9232&0.3544\\
				cyclists&0.0030&0.0738&0.9213&0.5382&0.0019&0.0488&0.9492&0.3457\\
				flock&0.0048&0.0641&0.9383&0.9270&0.0036&0.0375&0.9591&0.6179\\
				freeway&0.0013&0.3012&0.7787&0.5480&0.0028&0.1349&0.8394&0.4635\\
				hockey&0.0197&0.0716&0.9165&2.8430&0.0138&0.0527&0.9400&2.0359\\
				jump&0.0066&0.0746&0.9358&1.4309&0.0041&0.0464&0.9603&0.8892\\
				landing&0.0007&0.0653&0.9245&0.1278&0.0006&0.0559&0.9388&0.1031\\
				ocean&0.0012&0.1014&0.8931&0.2253&0.0010&0.0607&0.9243&0.1615\\
				peds&0.0048&0.0955&0.8776&0.7499&0.0038&0.0907&0.8942&0.6402\\
				rain&0.0029&0.1180&0.9174&1.0490&0.0025&0.0558&0.9534&0.5881\\
				skiing&0.0022&0.0846&0.9171&0.4338&0.0018&0.0586&0.9385&0.3251\\
				surfers&0.0015&0.1147&0.8887&0.2990&0.0012&0.0764&0.9173&0.2232\\
				surf&0.0008&0.2941&0.7307&0.1778&0.0005&0.2321&0.7968&0.1343\\
				traffic&0.0017&0.0899&0.9070&0.3186&0.0015&0.0566&0.9301&0.2427\\
				zodiac&0.0003&0.0735&0.8988&0.0429&0.0002&0.0815&0.9086&0.0383\\
				\hline
				\rowstyle{\bfseries}FgSegNet\_v2&0.0033&0.1086&0.8945&0.6136&0.0025&0.0762&0.9203&0.4405\\
				\hline
				\rowstyle{\bfseries}FgSegNet\_S \cite{lim2018foreground}&0.0058&0.0559&0.8822&0.7052&0.0039&0.0544&0.9139&0.5024\\
				\rowstyle{\bfseries}FgSegNet\_M \cite{lim2018foreground}&0.0037&0.0904&0.8948&0.6260&0.0027&0.0714&0.9203&0.4637\\
				\hline
			\end{tabular}
		\end{table*}
	
\begin{figure*}[!h]
			\centering
			\includegraphics[width=0.98\textwidth]{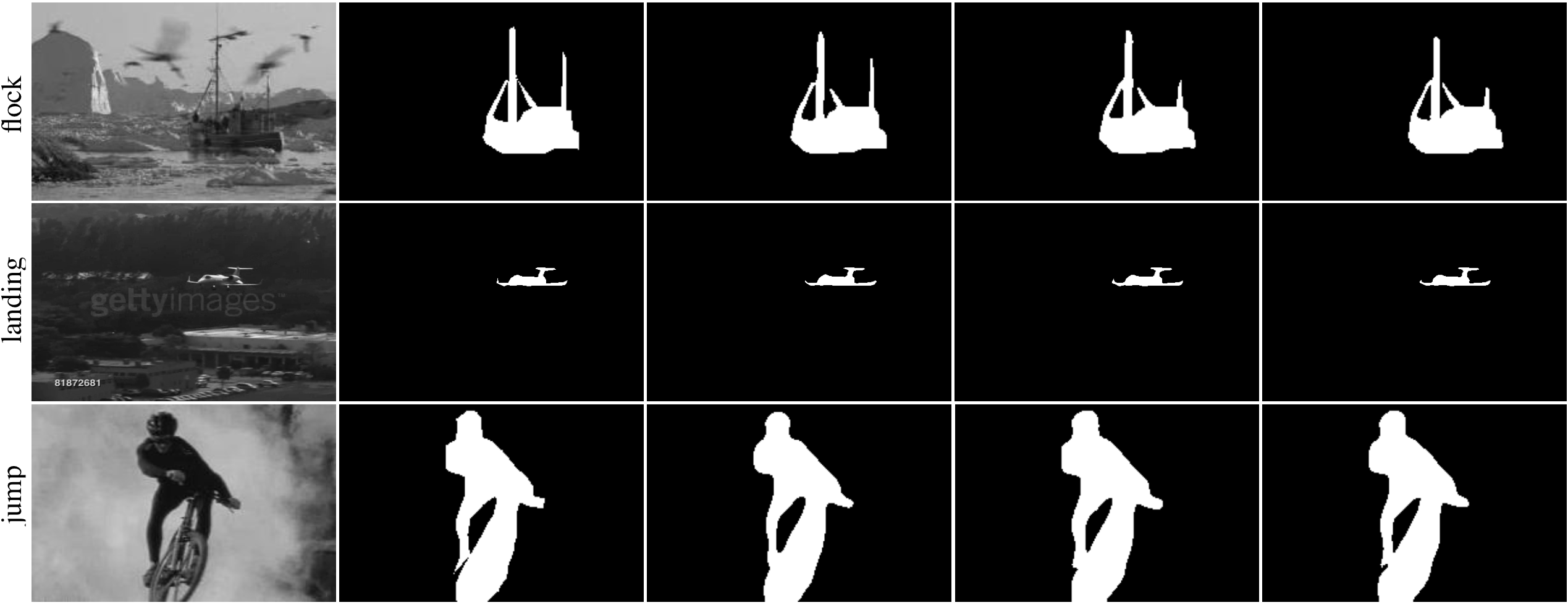}
			\caption{A comparison on UCSD dataset. Each column shows raw images, the ground-truths, our segmentation results, FgSegNet\_S and FgSegNet\_M results, respectively.}
			\label{fig:fig8}
		\end{figure*}

		Similar to \cite{lim2018foreground}, we further evaluate our method on UCSD Background Subtraction dataset \cite{mahadevan2010spatiotemporal}, which contains 18 video sequences with ground-truth labels. This dataset contains highly dynamic backgrounds, which are extremely challenging, and the number of frames are relatively small compared to CDnet2014 and SBI2015 dataset. We use the same training/testing splits provided by \cite{lim2018foreground}, where, first, the 20\% split is used for training, i.e. $t_n\in[3, 23]$, and 80\% for testing; second, 50\% is used for training, i.e. $t_n\in[7, 56]$, and remaining 50\% for testing. \textit{Test results} are depicted in Table \ref{table:6}. Although there are very small numbers of \textit{training+validation} frames, we obtain an average F-Measure of 0.8945 in case of 20\% split and 0.9203 in case of 50\% split. Our method produces comparable results to the previous methods, while PWC decreases remarkably compared to the previous methods. We also depict some segmentation results in Fig. \ref{fig:fig8}.
	
\section{Conclusion}
\label{sec:6}
	In this work, we propose a robust encoder-decoder network, which can be trained end-to-end in a supervised manner. Our network is simple, yet can learn accurate foreground segmentation by using a few training examples, which alleviates the need of ground-truths labeling burden. We improve the original FPM module by fusing multiple scale features inside FPM module; resulting in a robust module against camera motion, which can alleviate the need for training the network with multi-scale inputs. We further propose a simple decoder, which can help improving the performance. Our method neither requires any post-processing to refine the segmentation results nor temporal data into consideration. The experimental results reveal that our network outperforms existing state-of-the-art methods in several benchmarks. As a future work, we plan to incorporate temporal data and redesign a method, which can learn from very small number of examples.
	
\section*{Acknowledgements}
	We would like to thank CDnet2014 benchmark \cite{wang2014cdnet} for making the segmentation masks of all methods publicly available, which allowed us to perform different types of comparisons. And we also thank the authors in \cite{WANG201766} who made their training frames publicly available for follow-up researchers.

\bibliographystyle{ieeetr}
\footnotesize \bibliography{fgsegnet_v2_arxiv}
\end{document}